\documentclass[journal]{IEEEtran}
\usepackage{float}
\usepackage{subfig}
\usepackage{caption}
\usepackage{algorithm}
\usepackage{algpseudocode, balance}

\ifCLASSINFOpdf
  \usepackage[pdftex]{graphicx}
  \graphicspath{{./}}
  \DeclareGraphicsExtensions{.pdf,.jpeg,.png}
  \usepackage[space]{grffile}
\else
  \usepackage[dvips]{graphicx}
  \graphicspath{{./eps/}}
  \DeclareGraphicsExtensions{.eps}
\fi
\usepackage{epstopdf}
\usepackage{epsfig}
\usepackage[cmex10]{amsmath}

\begin{document}
\title{ \sc Diffusion Least Mean Square: Simulations}

\author{Jonathan~Gelati\textsuperscript{1}, Sithan~Kanna\textsuperscript{2}\\
\textsuperscript{1}jonathan@softwareengineer.it\\
\textsuperscript{2}ssk08@ic.ac.uk
} % stops a space

\markboth{Gelati et. al., Diffusion Least Mean Square: Simulations}%
{}



\maketitle

\begin{abstract}
In this technical report we analyse the performance of diffusion strategies applied to the Least-Mean-Square adaptive filter. We configure a network of cooperative agents running adaptive filters and discuss their behaviour when compared with a non-cooperative agent which represents the average of the network. The analysis provides conditions under which diversity in the filter parameters is beneficial in terms of convergence and stability. Simulations drive and support the analysis.
\end{abstract}


% \fi
\IEEEpeerreviewmaketitle

\section{Introduction}
Machine learning (ML) is the area of artificial intelligence which studies how a software application can learn by repeated training \cite{mitchell97}. In ML, software applications are not systematically programmed step by step to a particular purpose, but they are instead able to evaluate data instances and generalize their own behaviour in order to perform on new unseen data. In signal processing, ML algorithms are called \emph{adaptive filters} and are used to extract an estimate of the desired signal when some parameters of the target signal are not known in advance. Adaptive filters are able to refine their update strategy by assessing the error at each time instant and can adapt to changing conditions over time. A widely used adaptive filter is the Least-Mean-Square (LMS) which aims at minimizing the squared difference between the desired and estimated signals. 

In modern information systems, it is frequent to deal with settings where it is not feasible to process the amount of data in a timely fashion or to collect them in a single place given they are spread across many sources. The limiting factor is having a single centralized computational centre which is capable to cope with the computational and communication workload. To face these challenges, ML naturally evolves into distributed ML (DML) where we use a network of nodes, typically organized in neighbourhoods where each neighbourhood uses a \emph{diffusion} adaptation strategy \cite{lopes07}: a node executes an ML algorithm, cooperates with others by sharing its estimations and combines the estimations of its neighbourhood using weighting coefficients. 

In this report, our focus will be on diffusion adaptive networks where nodes use the LMS adaptive filter for signal processing. We define a diffusion adaptive network as configured by a set of parameters: the initial vector parameter, the learning rate, the trust coefficients and the input and noise mean and standard deviations. To study how each parameter affects the network performance in terms of speed of convergence and stability, we consider the behaviour of the cooperative and non-cooperative nodes with equal configuration. We then compare the estimations of the cooperative nodes with the average of the estimations of the non-cooperative nodes. It turns out that introducing diversity in the network can improve convergence. The most evident gain in performance is when nodes have different learning rates: having two nodes with two different rates is more beneficial than having a single node with a learning rate which is the average of the learning rates from its distributed counterpart. Different initial vector parameters are in general merged after the first few iterations, after which the estimations of all nodes are equal. The number of iterations by which the estimations start to overlap is controlled by the trust coefficients: more selfish nodes require more iterations before the overlapping happens. In the case nodes perceive signal with different degrees of noise, the nodes which perceive the most noisy signals perform a weighted sum of estimations computed by nodes reading less noisy signals.    

The rest of the technical report is organized as follows: in Section \ref{sec:background} we will give the mathematical background about the LMS adaptive filter. In Section \ref{sec:model} we will define the agent model we use to study our adaptive network and detail the diffusion strategy we will use in our simulations. In Section \ref{sec:comparison} we will analyse how each parameter affects the behaviour of agents and discuss the results of experimental simulations. Finally in Section \ref{sec:conclusion} we will summarize the configurations of our network of agents which show that diffusion LMS can indeed perform better than LMS.
 
\section{Background}\label{sec:background}
In the following we use \textbf{X} to indicate the matrix of features with size $M$ x $L$, where $M$ is the number of features and $L$ the number of time instants. We use \textbf{y} for the vector of measured signals with size $L$ and \textbf{w} for the vector of parameters with size $M$.  The positive scalar $i$ is used to denote time instants.
 
\subsection{Gradient descent}
Given a function $J(i)$ defined on  variables \textbf{X} and with parameters \textbf{w}, the gradient descent algorithm applied to $J(i)$  possibly finds the values of the vector parameter \textbf{w} which minimize the value of $J(i)$. The algorithm iteratively computes the value of \textbf{w} at time $i$ as follows:
\begin{equation} \label{eq:Wupdate}
\mathbf{w}(i) = \mathbf{w}(i-1) - \mu \frac{\partial J(\mathbf{w}(i-1))}{\partial \mathbf{w}} 
\end{equation}
The parameter $\mu$ is called \emph{learning rate} and is used to control the step size at each iteration. For smaller values of $\mu$ the algorithm will converge more slowly to the values of the vector \textbf{w} which minimize $J(i)$. As we assign larger values to $\mu$ the algorithm converges more quickly and possibly will diverge, showing that the value of $\mu$ is too large for the case at hand.
$J(i)$ is usually a cost function related to some given function on variable $\mathbf{X}$ having as parameter vector $\mathbf{w}$, denoted as $h_w (\mathbf{X})$. A popular example of cost function is the sum of squares where $h_w$ is the linear function: 

\begin{equation} \label{eq:gdcostfunction}
J(i) = \frac{1}{2L}\sum\limits_{k=1}^L (y(k) - \mathbf{w}(i)^T \mathbf{x}(k))^2
\end{equation}
where $\mathbf{x}(k)$ is a row of $\mathbf{X}$.
Applying the gradient descent algorithm to $J(i)$ means finding the parameter vector \textbf{w} which minimizes the error between the predicted and the actual target values, leading to the following update strategy:
\begin{equation} \label{eq:gdupdatestrategy}
\mathbf{w}(i) =\mathbf{w}(i-1) + \mu \frac{1}{L} \sum\limits_{k=1}^L  (y(k) - \mathbf{w}(i-1)^T \mathbf{x}(k))) \mathbf{x}(k)
\end{equation}
where $\mu$ includes the constant scalar resulting from the derivative operation.

\subsection{LMS adaptive filter}
Equation (\ref{eq:gdcostfunction}) assumes that we know in advance all of the values of $\mathbf{x}(i)$. This is not the case for online or real-time applications. For this class of applications, we can use an instantaneous gradient, that results in the weight update 
\begin{equation}
\label{eq:instantaneous_gradient}
\mathbf{w}(i) = \mathbf{w}(i-1) + \mu  (y(i) - \mathbf{w}(i-1)^T \mathbf{x}(i))) \mathbf{x}(i).
\end{equation}
Such an adaptive filter is called \emph{stochastic gradient adaptive filter} as it makes use of the instantaneous gradient which according to  \cite{mathews03} ``\emph{is an unbiased estimate of the true gradient. Since the step parameter is chosen to be a small value, any errors introduced by the instantaneous gradient are averaged over several iterations, and thus the performance loss incurred by its approximation is relatively small}''. 

\section{Diffusion LMS}\label{sec:model}
Diffusion LMS (DLMS) is used in settings where more filters are simultaneously run to estimate the same optimal vector parameter. It extends LMS by introducing an additional step where the estimations of the filters are combined. The combination step may occur before or after the execution of the update strategy reported in Equation (\ref{eq:instantaneous_gradient}).
 
\subsection{Model}
For the purpose of our investigation, we model a distributed machine learning environment as a set of $N$ agents similarly to \cite{sayed13}. An agent is an independent computational unit, which perceives the input signal with a certain degree of noise and iteratively applies the gradient descent algorithm to compute the parameter vector $\mathbf{w}$. The goal of such an agent is to find the parameter vector $\mathbf{w_{opt}}$ which minimizes the cost function $J(i)$.

While being computationally independent, an agent $a$ may share information about its estimates with other agents. From a topological viewpoint, an agent $a$ belonging to a network exchanges information with a subset of other agents belonging to the same network. This subset is defined as the neighbourhood of agent $a$ denoted as $\mathcal{N}_a$ and it is here intended as ``physical" neighbourhood: if agent $a$ is a neighbour of agent $b$ then opposite also  holds. In general, a network of agents is an undirected graph where agents may change the neighbourhoods they belong to over time \cite{sayed12}. Another aspect of the dynamics of a network is related to how much trust has an agent $a$ for the information its neighbour $b$ shares with it. This is a directional property from agent $a$ to agent $b$ and it is usually indicated as the scalar $s_{ab}$. Note that $s_{ab}$ and $s_{ba}$ need not hold the same value.   
  
Independently from the strategies adopted to diffuse information among them, agents can share the following data:
\begin{itemize}
  \item the estimated parameter vector $\mathbf{w}$ computed at each gradient descent iteration;
  \item the gradient approximation at each gradient descent iteration;
  \item the history of the above information.
\end{itemize} 

\subsection{Diffusion strategy}
To the purpose of our analysis we have built a simulation software to execute multi-agent systems where each agent processes the input signal using a gradient descent algorithm. In our multi-agent system a computational iteration is composed by two execution steps: first we run the gradient descent algorithm for each agent and then when all agents have completed their computations the agents share their estimated vector parameter $\mathbf{w}$ with their neighbours by applying a weighted sum based on trust coefficients. This strategy is also called combine-then-adapt (CTA) in  \cite{sayed13} and is summarized in Algorithm \ref{alg:CTA}. 

\begin{algorithm} 
\caption{DLMS algorithm using a CTA diffusion strategy to estimate the vector parameter $\mathbf{w}$}
\label{alg:CTA}
	\begin{algorithmic}
		\For {time instant $i = \left\{ {1, \ldots, L}\right\}$}
			\For {agent $a = \left\{ {1, \ldots, N}\right\}$}
				\State \textbf{Input data:} $\mathbf{x}_{a}(i)$, $\mathbf{w}_{a}(0)$
				\State \textbf{Desired signal:} $y_{a}(i)$
				\State $\mathbf{\boldsymbol \psi}_a(i) = \sum\limits_{b \in {N}_a}   s_{ab} \mathbf{w}_b(i-1)$
				\State $e_a(i) =  y_a(i) -  \mathbf{\boldsymbol \psi}_a(i)^T \mathbf{x}_a(i)$
				\State $\mathbf{w}_a(i) = \mathbf{\boldsymbol \psi}_a(i) + \mu  e_a(i) \mathbf{x}_a(i)$
                 		\EndFor
                 	\EndFor
	\end{algorithmic}
\end{algorithm}

From Equation \ref{eq:instantaneous_gradient} and Algorithm \ref{alg:CTA} we identified the following set of parameters as characterizing the configuration of a single agent in a DLMS network:
\begin{itemize}
  \item the learning rate $\mu$;
  \item the initial value of the vector parameter $\mathbf{w}$;
  \item the trust coefficients for each neighbour;
  \item the mean and standard deviation of the perceived input signal. 
\end{itemize} 

\section{How DLMS outperforms the average filter}\label{sec:comparison}
Our goal is to analyse whether having two agents cooperating is beneficial in terms of how fast they converge - the number of iterations needed to get close to $\mathbf{w}_{opt}$  - and in terms of the variance of the error. In our experiments we consider the scenario of two cooperative agents. We configure agent $a$ and agent $b$ to be two cooperative agents, agent $c$ and agent $d$ to be non-cooperative agents having the same configuration of agents $a$ and $b$ respectively and finally agent $e$ to be an agent which does not run gradient descent, does not perceive any signal and merely averages the estimations of the non-cooperative agents $c$ and $d$. This is done in order to compare the cooperative agents with their standalone counterparts and to compare how they behave with respect to an agent merely averaging the estimations of the two standalone agents.
In our simulations agents have all the following common configuration:
\begin{itemize}
  \item the function $h_w(\mathbf{x}(i)) = \mathbf{w}(i-1)^T \mathbf{x}(i)$, where vectors are one-dimensional;
  \item the update strategy in Equation (\ref{eq:instantaneous_gradient});
 \item the perceived signal given by:
\begin{equation} \label{eq:target_linear_generation}
y(i) = \mathbf{w}_{opt}^T \mathbf{x}(i) + q(i)
\end{equation}
where $\mathbf{x}(i)$ is the input and $q(i)$ is the measurement noise. In our simulations we used the Box Muller transformation \cite{boxmuller58} to randomly generate the two signals with a given mean and standard deviation. Note that in our experiments cooperative agents always perceive uncorrelated inputs.
\end{itemize}

To be informative while discussing the results of our experimental, runs we will make use of an analogy to describe the agent behaviours. Agents trying to estimate the optimal vector parameter are seen like people being in a street and try to walk towards a target position. Each person can have her initial position (initial vector parameter), its step size (the learning rate) and sight (signal perception). Analogies will be highlighted in italics.
 
\subsection{Diversity in initial vector leads to an average behaviour}
The simplest diversity we can think of is having agents with the same configuration but the initial vector parameter, which determines the starting point of an agent and how far the agent is from the optimal vector parameter. In general, in the network of agents there will be agents closer to the optimal  and some others further from it. Given that agents share data at each iteration, we expect that after the first time they average their estimations they behave exactly the same and as the average agent. 

For our simulations we used the configuration detailed in TABLE \ref{tab:twoagents_par_w0}.

\begin{center}
\begin{tabular}{ | l | c | c | c | c | c |}
\hline
 & $\mu$ & $\mathbf{w}_0$ & Trust coeff. & Input SD & Noise SD \\ \hline
$a$ & 0.5 & 0 & $s_{aa}$: 0.5 $s_{ab}$: 0.5  & 0.09 & 0.03\\ \hline
$b$ & 0.5 & 1 & $s_{bb}$: 0.5 $s_{ba}$: 0.5  & 0.09 & 0.03\\ \hline
$c$ & \multicolumn{5}{c|} {agent $a$ with no cooperation}  \\ \hline
$d$ & \multicolumn{5}{c|} {agent $b$ with no cooperation} \\ \hline
$e$ & \multicolumn{5}{c|} {The average of the non-cooperative agents $c$ and $d$}  \\ \hline
%\captionof{table}{Values of parameters for each agent} \label{tab:twoagents_par_w0}
\end{tabular}
\captionof{table}{Values of parameters for each agent.} \label{tab:twoagents_par_w0}
\end{center}

\begin{figure}[H]
\centering
\includegraphics[scale=0.5,clip, trim =10mm 75mm 10mm 10mm]{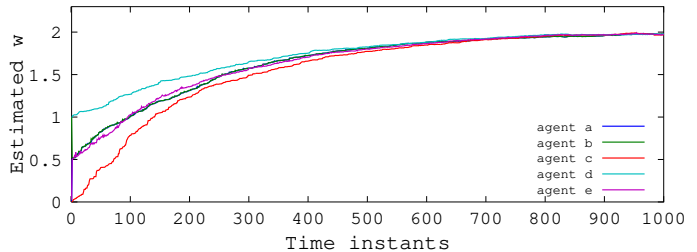}
\caption{Agents differing by the initial vector parameter.}  
  \label{fig:twoagents_w0}
\end{figure}

In Fig. \ref{fig:twoagents_w0} we see how having two agents with a different initial vector parameter $\mathbf{w}_0$ impacts on convergence. We have a topology with two agents, equally trusting each other ($s_{aa}$ = $s_{ab}$ = $s_{bb}$ = $s_{ba}$ = 0.5). For this particular case, the function to share estimations reduces to the average function, being the balanced weighted sum of $\mathbf{w}$ multiplied by the trust coefficients. 

What we notice is that averaging estimations has an impact especially in the second iteration, i.e. the first iteration where agents take into consideration the shared data. From the second iteration on, the estimations of the cooperative agents $a$ and $b$ get much closer, the estimation of agent $b$ levelling down and the one of agent $a$ levelling up. From that moment they behave similarly to the standalone agent $e$ which has an initial vector parameter equal to the average of their initial vector parameters. We see how the plots of $\mathbf{w}$ for agents $a$ and $b$ follow with a delay the one of agent $c$.  

\emph{We can think the agents as two people in the same street at different locations. After the first step, they share where they are and before going further they decide to get closer, choosing the middle point between the two. From that moment on they proceed with the same step size along the street, following a lonely person who initially started her walk at the middle point.} 

\subsection{Heterogeneous learning rates boost diffusion convergence rate}\label{subsec:learningrate}
Consider the case where agents in the network differ by the learning rate. In general, a standalone agent with a higher learning rate converges at a faster pace. On the other hand, for a network with two acquainted agents the following holds: 
\begin{equation} \label{eq:cooperative_agents}
\begin{split}
\mathbf{\boldsymbol \psi}_{a}(i) = \mathbf{w}_{a}(i - 1) + s_{ab} (\mathbf{w}_{b}(i - 1) - \mathbf{w}_{a}(i - 1))\\
 \mathbf{\boldsymbol \psi}_{b}(i) = \mathbf{w}_{b}(i - 1) + s_{ba} (\mathbf{w}_{a}(i -1) - \mathbf{w}_{b}(i - 1))
\end{split}
\end{equation}
The original estimation of an agent is adapted using the difference between its estimation and the one of the neighbour. Thus, an agent having a neighbour which learns faster than it will see its estimation improved toward the optimal value. No matter what the trust coefficients are, the weighted sum estimation will be better than the original estimations of the slower learning agents. The net effect is that the faster learning agent has a beneficial influence over the slower agent. To gain a better insight of this case we set up and run the following simulations.   
  
First, we observed the behaviour of two cooperative agents having different learning rate. The configuration is detailed in TABLE \ref{tab:twoagents_par_mu}.
\begin{center}
\captionof{table}{Values of parameters for each agent.} \label{tab:twoagents_par_mu}
\begin{tabular}{ | l | c | c | c | c | c |}
\hline
 & $\mu$ & $\mathbf{w}_0$ & Trust coeff. & Input SD & Noise SD \\ \hline
$a$ & 0.2 & 0 & $s_{aa}$: 0.5 $s_{ab}$: 0.5  & 0.09 & 0.03\\ \hline
$b$ & 0.8 & 0 & $s_{bb}$: 0.5 $s_{ba}$: 0.5  & 0.09 & 0.03\\ \hline
$c$ & \multicolumn{5}{c|} {agent $a$ with no cooperation}  \\ \hline
$d$ & \multicolumn{5}{c|} {agent $b$ with no cooperation} \\ \hline
$e$ & \multicolumn{5}{c|} {The average of the non-cooperative agents $c$ and $d$}  \\ \hline
 \end{tabular}
\end{center}

\begin{figure}[H]
\centering
\includegraphics[scale=0.5,clip, trim =10mm 75mm 10mm 10mm]{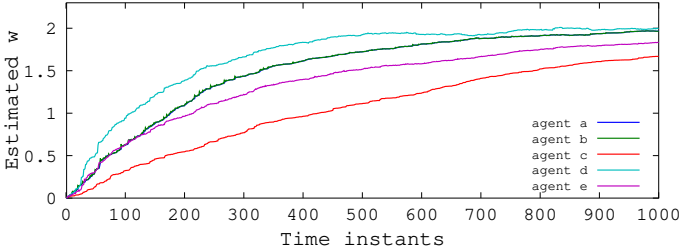}
\caption{Agents differing by the learning rate.}  
  \label{fig:twoagents_mu}
\end{figure}

Again we notice that in Fig. \ref{fig:twoagents_mu} the first two iterations are used to balance the estimations of agents $a$ and $b$. Differently from the previous case, after each step a small gap is produced in the estimations of the two agents because of their different $\mu$ values. The computed $\mathbf{w}$ is different for the two agents even after having performed the same weighted sum of the combined estimations. 

\emph{We can think the agents as three people $a$, $b$ and $e$ in the same street and at the same location. Person $a$ is tall and has a step size bigger than person $b$. Person $e$ has a step size which is exactly the average of the other two. The goal of all three people is to arrive at the same target location. The first two people $a$ and $b$ know each other while the third one is a lonely person. After each step $a$ has gone further than person $b$. Given they are acquainted they share their positions and both move to the middle point between the two. As the steps of $a$ and $b$ may also differ in direction, the two people go along a somewhat twisty path. In the case both accidentally choose a better path than the one taken by person $e$ they may even pass by the third person who has a step size average of the two. As the target location gets closer, all three have a better sight of the target location and are finally very close to each other. }

In order to have a better evidence of what we observed, we analysed the behaviour of two cooperative agents having different initial vector parameter and learning rate. The configuration is detailed in TABLE \ref{tab:par_w0_mu}.

\begin{center}
\captionof{table}{Values of parameters for each agent.} \label{tab:par_w0_mu}
\begin{tabular}{ | l | c | c | c | c | c |}
\hline
 & $\mu$ & $\mathbf{w}_0$ & Trust coeff. & Input SD & Noise SD \\ \hline
$a$ & 0.2 & 0 & $s_{aa}$: 0.5 $s_{ab}$: 0.5  & 0.09 & 0.03\\ \hline
$b$ & 0.8 & 1 & $s_{bb}$: 0.5 $s_{ba}$: 0.5  & 0.09 & 0.03\\ \hline
$c$ & \multicolumn{5}{c|} {agent $a$ with no cooperation}  \\ \hline
$d$ & \multicolumn{5}{c|} {agent $b$ with no cooperation} \\ \hline
$e$ & \multicolumn{5}{c|} {The average of the non-cooperative agents $c$ and $d$}  \\ \hline
 \end{tabular}
\end{center}
\begin{figure}[!htbp]
\centering
\includegraphics[scale=0.5,clip, trim =10mm 75mm 10mm 10mm]{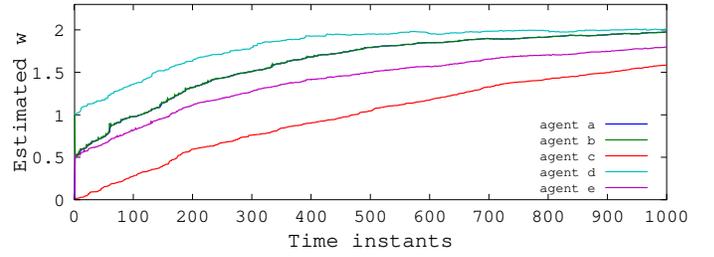}
\caption{Agent $a$ is better positioned and faster learning than agent $b$.}
 \label{fig:twoagents_w0_mu}
\end{figure}
In Fig. \ref{fig:twoagents_w0_mu} the agent which has the largest learning rate is now also better placed with respect to the optimal vector parameter.  An agent which such features exerts a positive influence over the other agent. During the combination step of the diffusion algorithm agent $a$ will see its estimated $\mathbf{w}$ improve while agent $b$ will see its previously computed estimation to get further from the optimal value. 
The intuition is that at time instant $i$ the agent with the lowest learning rate uses a value of $\boldsymbol \psi_i$ that is better of the $\mathbf{w}_i$ computed at the previous step $i-1$. Analogously  the agent with the highest learning rate sees its estimate $\mathbf{w}_i$ worsen.  In our experiment agent $a$ and agent $b$ still go hand in hand after the second iteration and their estimations more closely follow the one of the standalone agent which starts at a better position and has a higher learning rate. This is consistent with what we observed in the previous section for the cases where only $\mathbf{w}_0$ or $\mu$ changed: having an agent which is faster and better placed makes all network agents converge more quickly. 

Next we will observe the case where the agent with the lower learning rate is given an initial value that is closer to the optimal weight. 

\begin{center}
\captionof{table}{Values of parameters for each agent.} \label{tab:par_w0_mu_2}
\begin{tabular}{ | l | c | c | c | c | c |}
\hline
 & $\mu$ & $\mathbf{w}_0$ & Trust coeff. & Input SD & Noise SD \\ \hline
$a$ & 0.8 & 0 & $s_{aa}$: 0.5 $s_{ab}$: 0.5  & 0.09 & 0.03\\ \hline
$b$ & 0.2 & 1 & $s_{bb}$: 0.5 $s_{ba}$: 0.5  & 0.09 & 0.03\\ \hline
$c$ & \multicolumn{5}{c|} {agent $a$ with no cooperation}  \\ \hline
$d$ & \multicolumn{5}{c|} {agent $b$ with no cooperation} \\ \hline
$e$ & \multicolumn{5}{c|} {The average of the non-cooperative agents $c$ and $d$}  \\ \hline
 \end{tabular}
\end{center}
\begin{figure}[!htbp]
\centering
\includegraphics[scale=0.5,clip, trim =10mm 75mm 10mm 10mm]{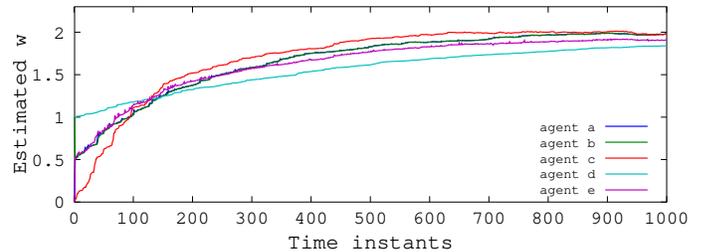}
\caption{Agent $a$ is better positioned and slower learning than agent $b$}
  \label{fig:twoagents_w0_mu_2}
\end{figure}
If we look at the standalone agents in Fig. \ref{fig:twoagents_w0_mu_2}, we note that even if agent $c$ starts from a vector parameter further from the optimal value than agent $d$, agent $c$ passes by agent $d$ around iteration 150.  From that iteration we have the same situation as in the previous case: the agent with the estimation closer to the optimal is also the faster learning agent. This explains why the cooperative agents start to perform better than the averaging agent $e$ from around iteration 450. We conclude that in both cases cooperation is an advantage in terms of convergence rate when there is diversity in learning rates.  

\subsection{Non-symmetric trust coefficients lead to a delayed average behaviour}
\begin{figure}[!htbp]
\centering
\includegraphics[scale=0.5,clip, trim =10mm 0mm 10mm 10mm]{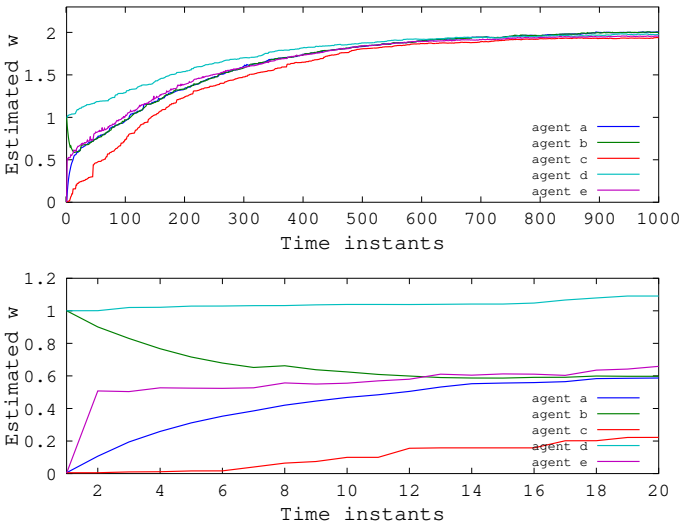}
\caption{\emph{Top panel}: Agents differing by trust coefficients. \emph{Bottom panel}: Weight trajectory for the first 20 iterations.}  
  \label{fig:twoagents_w0_s}
\end{figure}

Consider the case of two cooperative agents which are selfish, trusting their estimation more than they trust the one of the neighbour. This is to say that the influence exerted by an agent on its neighbour weakens. Fig. \ref{fig:twoagents_w0_s} shows that a delay is introduced before the agents start to behave like the average agent. We used the configuration detailed in TABLE \ref{tab:par_w0_s}.

\begin{center}
\captionof{table}{Values of parameters for each agent} \label{tab:par_w0_s}
\begin{tabular}{ | l | c | c | c | c | c |}
\hline
 & $\mu$ & $\mathbf{w}_0$ & Trust coeff. & Input SD & Noise SD \\ \hline
$a$ & 0.5 & 0 & $s_{aa}$: 0.9 $s_{ab}$: 0.1  & 0.09 & 0.03\\ \hline
$b$ & 0.5 & 1 & $s_{bb}$: 0.9 $s_{ba}$: 0.1  & 0.09 & 0.03\\ \hline
$c$ & \multicolumn{5}{c|} {agent $a$ with no cooperation}  \\ \hline
$d$ & \multicolumn{5}{c|} {agent $b$ with no cooperation} \\ \hline
$e$ & \multicolumn{5}{c|} {The average of the non-cooperative agents $c$ and $d$}  \\ \hline
 \end{tabular}
\end{center}
\emph{We can think the agents as two people in the same street at the same location and with the same step size. As they do not fully trust the choices of the other person their paths take longer to get close enough to overlap. When the overlapping happens they will proceed hand in hand as they have the same step size and they will basically follow pretty much the path of a person which just averages the positions at each iteration.}

\subsection{Variance in signal perception stabilizes the network}
We now consider the diversity produced by a different perception of the signal. The noise of the signal might be due to a number of factors such as the location of an agent or the sensors it uses. At each iteration the agent performs a weighted sum of expectations computed starting from signals with different noise variance.  We expect that this operation results in a more steady estimation. This is due to the fact that the weighted sum of two numbers, $z = s_{ab}  x + s_{ba}y$, where weights are less than one, $s_{ij} \in [0,1]$, has a variance lower than or equal to the highest variance of the two numbers: 
\begin{equation} \label{eq:weighted_sum_variance}
\begin{split}
\mathrm{Var}[z] = s_{ab}^2  \mathrm{Var}[x] + s_{ba}^2   \mathrm{Var}[y] + 2 s_{ab} s_{ba} \mathrm{Cov}[x,y]  
\end{split}
\end{equation}

Note that in equation (\ref{eq:weighted_sum_variance})  $\mathrm{Cov}[x,y]$ is zero if $x$ and $y$ are independent from each other as we assumed our input to be. To verify the expected behaviour, we ran simulations with the configuration detailed in TABLE \ref{tab:par_w0_noise}.  
\begin{center}
\captionof{table}{Values of parameters for each agent.} \label{tab:par_w0_noise}
\begin{tabular}{ | l | c | c | c | c | c |}
\hline
 & $\mu$ & $\mathbf{w}_0$ & Trust coeff. & Input SD & Noise SD \\ \hline
$a$ & 0.5 & 0 & $s_{aa}$: 0.5 $s_{ab}$: 0.5  & 0.09 & 0.01\\ \hline
$b$ & 0.5 & 1 & $s_{bb}$: 0.5 $s_{ba}$: 0.5  & 0.09 & 0.2\\ \hline
$c$ & \multicolumn{5}{c|} {agent $a$ with no cooperation}  \\ \hline
$d$ & \multicolumn{5}{c|} {agent $b$ with no cooperation} \\ \hline
$e$ & \multicolumn{5}{c|} {The average of the non-cooperative agents $c$ and $d$}  \\ \hline
 \end{tabular}
\end{center}
The fact that agents have a different initial vector parameter does not impinge on the estimation variance and only helps reading the plots.
  
\begin{figure}[!htbp]
\centering
\includegraphics[scale=0.5,clip, trim =10mm 75mm 10mm 10mm]{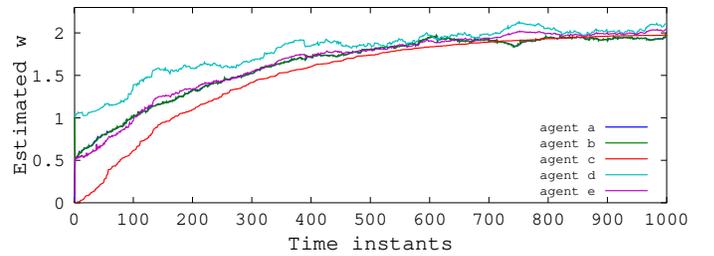}
\caption{Agents perceiving the signal with different variances}  
  \label{fig:twoagents_w0_noise}
\end{figure}
In Fig. \ref{fig:twoagents_w0_noise} the agents start from different initial vector parameters and as expected their estimations overlap after the second iteration. It is also important to highlight how the noise variance of the estimations of agent $a$ and agent $b$ is lower if compared with the noise variance of the standalone agent $d$ which has the largest standard deviation.  This means that a network of cooperative agent tends to flatten the effects of the agents which perceive more noisy signals.

\emph{We can think of two people in the street having different views of the path leading to their common target position. Sharing and averaging their positions at each time instant helps the blindest person to go on a less erroneous path.  The helping person is inevitably forced to alter its path to keep close to the helped person.} 

 \section{Conclusion}\label{sec:conclusion}
A network of agents executing DLMS filters and sharing estimations at each iteration can indeed perform better than the agent averaging an homogeneous adaptive network when there is diversity in the configuration of agents. Diversity is expressed by a different combination of parameters such as initial vector parameter, learning rate, trust coefficients or perceived signal. We can summarize the impact of each parameter as follows:
\begin{itemize}
  \item having agents with different initial vector parameters $\mathbf{w}_0$ determines which average configuration the network of agents behaves like after the second iteration;
  \item having agents with different $\mu$ values implies that the network of agents is composed by faster and slower agents which at each iteration reconcile their estimations. The net effect is a faster convergence compared to the behaviour of the average agent in a non-cooperative network;
  \item having agents which trust their own estimations more than they trust the estimations of other agents delays the time instant from which the estimations start to coincide and the network of agents starts to behave as a non-cooperative agent with an initial vector parameter which is the average of all initial vector parameters of the agents composing the network;
  \item having agents which perceive signal with different noise levels makes the network of agents helps stabilize the agents which perceive the more noisy signals.
\end{itemize} 
Finally, note that if the agents simultaneously differ for more parameters we get a combined effect on the behaviour of the filters.

\balance


\begin{thebibliography}{1}

\bibitem{mitchell97}
Mitchell, T., \emph{Machine Learning}, McGraw Hill. ISBN 0-07-042807-7, 1997.
\bibitem{lopes07}
C. Lopes and A. Sayed, \emph{Diffusion Least-Mean Squares Over Adaptive Networks}, in Proc. of the IEEE International Conference on Acoustics, Speech and Signal Processing, vol. 3, pp III-917 III-920, 2007. 
\bibitem{mathews03}
V John Mathews and Scott C Douglas, \emph{Adaptive filters}, Chapter 4 Stochastic Gradient Adaptive filters, 2003.
\bibitem{sayed13}
Ali H. Sayed, Sheng-Yuan Tu, Jianshu Chen, Xiaochuan Zhao and Zaid J. Towfic, \emph{Diffusion strategies for Adaptation and Learning over Networks}, IEEE Signal Processing Magazine \relax May 2013.
\bibitem{sayed12}
A. Sayed, \emph{Diffusion Adaptation over Networks}, CoRR, 2012.
\bibitem{boxmuller58}
G. E. P. Box and Mervin E. Muller, \emph{A Note on the Generation of Random Normal Deviates}, The Annals of Mathematical Statistics, Vol. 29, No. 2 pp. 610–611, 1958.
%\bibitem{cheng-tao06}
%Cheng-Tao Chu; Sang Kyun Kim, Yi-An Lin, YuanYuan Yu, Gary Bradski, Andrew Ng, and Kunle Olukotun, \emph{Map-Reduce for Machine Learning on Multicore}. NIPS, 2006.
%\bibitem{andrews00}
%Andrews Gregory R., \emph{Foundations of Multithreaded, Parallel, and Distributed Programming}, Addison–Wesley, ISBN 0-201-35752-6, 2000.
\end{thebibliography}
\end{document}